\documentclass[letterpaper, 10 pt, conference]{ieeeconf}  % Comment this line out if you need a4paper

\IEEEoverridecommandlockouts                              % This command is only needed if 
\title{\LARGE \bf
Navigation Explainability for Object Avoidance with LLMs for Assistive Robots
}

\author{Courtney Young, Yi Lian, J. Taery Kim, Sehoon Ha}

\usepackage{cite}
\usepackage{amsmath,amssymb,amsfonts}
\usepackage{algorithmic}
\usepackage{graphicx}
\usepackage{subcaption}
\usepackage{textcomp}
\usepackage{xcolor}
\usepackage{booktabs}
\usepackage{float}
\usepackage{ulem}
\usepackage{float}
\usepackage{tabularx}
\usepackage{dblfloatfix}
\usepackage{subcaption}
\usepackage{booktabs}
\usepackage{array} 
\usepackage{placeins}  
\let\labelindent\relax
\usepackage{enumitem}
\usepackage{tcolorbox}
\tcbuselibrary{skins, breakable}
\begin{document}

\newcommand{\cherry}[1]{\textcolor{blue}{[Cherry: #1]}}
\newcommand{\sehoon}[1]{\textcolor{magenta}{[Sehoon: #1]}}
\newcommand{\taery}[1]{\textcolor{orange}{[Taery: #1]}}
\newcommand{\courtney}[1]{\textcolor{red}{[Courtney: #1]}}

\maketitle
\thispagestyle{empty}
\pagestyle{empty}

%%%%%%%%%%%%%%%%%%%%%%%%%%%%%%%%%%%%%%%%%%%%%%%%%%%%%%%%%%%%%%%%%%%%%%%%%%%%%%%%
\begin{abstract}  

Blind and visually impaired individuals face uncertainty when navigating unfamiliar areas. Whether a cane or a guide dog is used for navigation, this may prove to be challenging. 
To mitigate this challenge, we present the Context-Aware Navigation Explanation framework that explains guide dog movements for obstacle avoidance and navigation using a large language model. 
This system integrates perceptual context with planned future trajectories to generate natural-language explanations of robot movement. 
We deploy the system on a Unitree Go2 platform in indoor navigation scenarios and evaluate how different explanation context configurations affect user trust and explanation satisfaction. Results show a strong positive relationship between explanation satisfaction and trust, and suggest that combining obstacle information with future trajectory leads to the most effective explanations. These findings highlight the potential of context-aware language explanations to improve transparency and trust in assistive robot navigation.

\end{abstract}

\section{INTRODUCTION}
% \begin{itemize}
%     \item Motivation: why transparency/explainability matters in assistive robotics
%     \begin{itemize}
%         \item Links to user trust and usability
%     \end{itemize}
%     \item Gap in literature: lack of natural language explanations for assistive technology
%     \item Hypothesis: The trust in the system is expected to vary based on the comments generated by the LLM due to different combinations of parameters.
%     \item My approach: LLM-generated navigation guidance
%     \item Contributions:
%     \begin{itemize}
%         \item No finetuning for LLM or YOLO
%         \item Investigation of LLM’s ability to create quality explanation based on amount and type of data
%         \item Investigation of users’ preferences for different LLM generated comments linking to overall robot trust
%     \end{itemize}
% \end{itemize}

Guide dogs can substantially improve the quality of life for blind and visually impaired (BVI) individuals \cite{rickly2021travelling, whitmarsh2005benefits}. However, high training costs and short service lives limit access to approximately 2\% of the BVI population \cite{nfb2019statistics}. To address this gap, researchers have developed guide robots in both wheeled and quadrupedal forms as potential assistive alternatives that can expand accessibility and complement animal guides \cite{guerreiro2019cabot, chen2023quadruped, xiao2021robotic, kim2023transforming, cai2024diverse}. Although these systems have shown promise, successful deployment in real-world environments depends not only on navigation capability, but also on whether users can understand the robot’s behavior, trust its decisions, and interact with it comfortably \cite{ambsdorf2022explain, kayukawa2022public, kim2025expectations, shih2025characteristics}.
% stange2020effects

Recent advances in Large Language Models (LLMs), Vision-Language Models (VLMs), and speech interfaces
% (TTS/STT)
create new opportunities for assistive navigation robots to interpret their surroundings and communicate their intent with natural language. This capability is especially important in BVI navigation, where users may not directly inspect the environment or infer why the robot is adjusting its path. However, it remains unclear which natural-language explanations are most effective in helping BVI users understand the robot’s behavior and build trust during navigation.

%Unlike animal guides, robots can combine multimodal sources of information, such as onboard sensor observations and spatial context, and express this understanding through natural language explanations \cite{liu2024dragon}. This capability is especially important in BVI navigation, where users may not be able to directly inspect the environment or infer why the robot is adjusting its path. However, it remains unclear what kinds of natural language explanations are most effective for helping BVI users understand the robot’s behavior and build trust during navigation.

%Robot set up here
\begin{figure}
    \centering
    \includegraphics[width=0.9\linewidth]{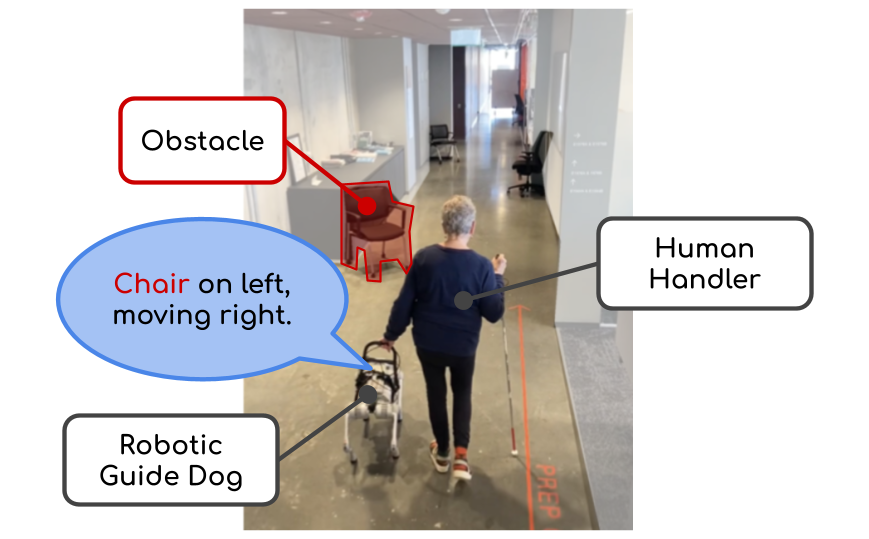}
    \caption{Overview of the Context-Aware Navigation Explanation system. The robot guide dog leads a BVI user while perceiving obstacles (e.g., a chair) and generating real-time natural language explanations (e.g., ``Chair on left, moving right'') based on scene context and planned motion.}
    % \vspace{-0.5em}
    \label{fig:teaser}
    \vspace{-1.5em}
\end{figure}

This study investigates how differences in robot guide dog navigation explanations affect user trust. We hypothesize that changes in the content and quality of LLM-generated comments, driven by different input context configurations, will lead to corresponding differences in trust. 
%For such explanations to be useful during navigation, they must reflect the robot’s behavior over time rather than a single isolated observation. 
% Because navigation unfolds over time, explanations based on a single isolated frame may miss the temporal context needed to interpret the robot’s ongoing behavior. Recent video-based VLMs further highlight the value of temporally grounded visual representations \cite{lin2024llava}. In assistive navigation, relying on fragmented context may increase user confusion and reduce trust in the robot.
%Single-frame context may miss the temporal information needed to explain ongoing motion, and recent video-based VLMs further highlight the value of temporally grounded visual representations \cite{lin2024llava}.

%To address this challenge, 
We propose a context-aware navigation explanation framework (Fig.~\ref{fig:teaser}), which generates real-time natural language explanations of a robot guide dog’s navigation behavior. This framework constructs a structured textual context by integrating detected obstacle information with future waypoints from a navigation planner over a short temporal window, then provides this context to an LLM to generate explanations of the robot’s current intent. By varying the amount and type of information included in this context, we evaluate how context design influences explanation satisfaction and overall system trust through a user study.
% Specifically, it constructs a structured textual context by integrating surrounding obstacle information extracted using an object detector with future waypoints from a navigation planner. This information is aggregated over a short temporal window and provided to an LLM to generate explanations that reflect the robot’s current intent. 

% \cite{stange2020effects, ambsdorf2022explain}.

%\sehoon{Summarize the results here...}X

%\sehoon{Finalize the contributions as bullet points...}X

Our results show a strong positive correlation between explanation satisfaction and trust, indicating that satisfying explanations are closely associated with increased user trust. We also observe that the condition combining obstacle information and future trajectory, with a moderate explanation frequency, generally received the highest ratings for both satisfaction and trust, outperforming both the no-future-trajectory and higher-frequency conditions. Item-level analysis further reveals that users particularly value explanations that are informative and complete to their goals.

Our contributions are as follows:
% \begin{itemize}
%     \item We propose a context-aware navigation explanation framework that integrates perception and future trajectory information to generate real-time natural language explanations for assistive navigation.
%     \item We investigate how different types and amounts of contextual information affect explanation quality and user trust through a controlled user study.
%     \item We show that combining obstacle awareness with future navigation intent improves both explanation satisfaction and perceived trust compared to limited-context explanations.
%     \item We provide insights into which aspects of explanations (e.g., completeness, usefulness, informativeness) most influence user trust in assistive navigation settings.
% \end{itemize}

\begin{itemize}
    \item We propose a context-aware navigation explanation framework that integrates perception and future trajectory information to generate real-time natural language explanations for assistive navigation.
    \item We investigate how different types and amounts of contextual information affect explanation satisfaction and user trust through a controlled user study.
    \item We show that combining obstacle awareness with future navigation intent improves both explanation satisfaction and perceived trust compared to limited-context explanations.
    \item We provide insights into which aspects of explanations (e.g., completeness, usefulness, informativeness) most influence user trust in assistive navigation settings.
\end{itemize}

\section{Related Work}

% \cherry{I will work on cutting Related Work} \sehoon{Yes, 58 seems to be too much. Even the half is fine.}

% Assistive guiding in human environments is safety-critical and interaction-heavy, requiring not only reliable navigation but also communication that helps users understand and anticipate robot motion. We review prior work on BVI guide robots and robot guide dogs, trust and communication in assistive HRI, and explainable navigation using multimodal foundation models to generate grounded language explanations.

Assistive guiding is a safety-critical task that requires reliable navigation and communication to help users understand and anticipate robot motion. We review prior work on BVI guide robots and robot guide dogs, trust barriers in assistive HRI, and the emergence of foundation models for explainable navigation. 

\subsection{Robot Guide Dogs and Assistive Guiding for BVI Users}

Robotic guidance for BVI users differs from generic mobile navigation: the system must lead safely in shared spaces while supporting a user’s situational awareness and interpretation of motion decisions, especially when behavior deviates from the user’s expectation \cite{kim2025expectations, hersh2010survey, shih2025characteristics}. 
% Early assistive navigation systems span wheeled guide robots and travel aids with tactile/haptic/audio feedback, including CaBot and classic devices such as the GuideCane, NavBelt, intelligent guide-sticks, MELDOG electrocutaneous communication, SWAN audio navigation, and spatial tactile interfaces \cite{guerreiro2019cabot, borenstein1997cane, shoval1998belt, kang2001stick, tachi1985meldog, wilson2007swan, yatani2012space}. Work on BVI navigation highlights feedback modality and cognitive load as central design constraints, shaping what information can be conveyed and when \cite{jiang2024haptic,shih2025characteristics}. Recent HRI work also examines adoption and deployment constraints in public settings and through participatory design \cite{azenkot2016enabling, kayukawa2022public, kayukawa2023museum, zhang2023boss, ranganeni2023levels, nanavati2019follow}.
Early assistive systems explored a range of feedback modalities, including haptic, audio, and tactile interfaces \cite{guerreiro2019cabot, borenstein1997cane},
% wilson2007swan, yatani2012space}
highlighting trade-offs between information richness and cognitive load \cite{jiang2024haptic, shih2025characteristics}. More recent work emphasizes real-world deployment challenges and participatory design with BVI users \cite{azenkot2016enabling, kayukawa2022public}.
% , ranganeni2023levels}.

% More recently, ``robot guide dog'' paradigms have expanded to include quadruped platforms that leverage the guide-dog metaphor and enable richer physical guidance (e.g., leash/tether interaction) and support guiding in a range of environments \cite{xiao2021robotic, chen2023quadruped, kim2023transforming, chuang2018trail, hamed2019hierarchical, wang2021navdog, hwang2022configuration, wang2023quadruped, hata2024spot, cai2024diverse, due2025walk, wei2014research}. These systems frame guiding as an integrated problem spanning navigation, physical interaction, safety, and intent communication \cite{wang2025cori, kim2023transforming, nanavati2019follow}. In particular, Kim et al. formalize guiding as a human--robot team task and introduce interaction modeling with an action-shielding mechanism to filter unsafe actions on a quadruped platform \cite{kim2023transforming}. User-centered studies further motivate richer communication: BVI users desire semantic feedback and explainability with personalization \cite{kim2025expectations, hersh2010survey, shih2025characteristics}, and analyses of handler--guide-dog partnerships suggest guide robots should relay pertinent information while managing cognitive load from excessive audio feedback \cite{hwang2024companions}.
Recent ``robot guide dog'' paradigms extend these ideas using mobile and quadruped platforms that support physical guidance and richer interaction \cite{xiao2021robotic, chen2023quadruped, kim2023transforming, cai2024diverse}. These systems frame guiding as an integrated problem spanning navigation, physical interaction, and safety \cite{kim2023transforming}. User-centered studies further highlight the importance of communication in assistive navigation: BVI users prefer semantic, personalized feedback that helps them interpret the robot’s behavior, while also emphasizing the need to manage cognitive load from excessive audio information \cite{kim2025expectations, hwang2024companions, shih2025characteristics}.

\subsection{Trust and Communication in Assistive HRI}

% Trust and communication are repeatedly identified as barriers to adoption for guide-dog robots. Handlers describe trust as emerging through experience and reliable behavior, and express uncertainty about whether comparable trust can be established with a robotic guide \cite{hwang2024companions, hersh2010survey}. Design studies with BVI stakeholders likewise emphasize the need for clear, user-tailored communication and explicitly call out explainability as a desired capability \cite{kim2025expectations, azenkot2016enabling, shih2025characteristics, kayukawa2022public}. Related work also shows that perceived trust and acceptance depend on how control is shared and how guidance is communicated to users and surrounding stakeholders \cite{ranganeni2023levels, zhang2023boss, kayukawa2022public, nanavati2019follow}. Across HRI, explanations can improve how users make sense of robot behavior, and their effectiveness depends on explanation type and delivery \cite{taschdjian2020cross, ambsdorf2022explain, stange2022self, setchi2020explainable, dehkordi2021explainability, anjomshoae2019explainable}.
Trust and communication remain central barriers to adoption for assistive robots. Studies with BVI users emphasize that trust develops through consistent behavior and clear communication, and may be difficult to establish without sufficient transparency \cite{hwang2024companions, hersh2010survey}. Participatory design work similarly identifies explainability and user-tailored feedback as critical requirements \cite{kim2025expectations, azenkot2016enabling, shih2025characteristics}. Across human-robot interaction (HRI), explanations have been shown to improve users’ understanding of robot behavior, though their effectiveness depends on content and delivery \cite{ambsdorf2022explain, stange2022self}.
% , taschdjian2020cross}. 

% Recent work studies how robots communicate intent during physical human--robot interaction, emphasizing interfaces that remain usable under attention and sensory constraints \cite{wang2025cori}. Dialogue-based assistive navigation systems such as DRAGON provide semantic grounding and conversational assistance for BVI users \cite{liu2024dragon}, but an open challenge remains in explaining \emph{motion decisions} (e.g., turning or stopping) in real time when they arise from planning constraints and dynamic scenes. Prior work on legible motion shows that trajectories can communicate intent, but constraints can still yield surprising deviations; more broadly, navigation trust and perceived predictability are shaped by implicit motion cues, motivating complementary language-based explanations when motion alone is ambiguous \cite{dragan2013legibility, dragan2013legible, lian2025implicit}.

Recent work studies how robots communicate intent under sensory and attention constraints \cite{wang2025cori}. Dialogue-based assistive systems such as DRAGON show how natural language can support BVI users during navigation \cite{liu2024dragon}, but explaining motion decisions in real time remains an open challenge when they arise from planning constraints and dynamic scenes. At the same time, implicit communication through motion plays an important role: legible trajectories can convey intent, but may still lead to ambiguity when environmental constraints cause unexpected deviations \cite{dragan2013legibility, lian2025implicit}. This motivates combining motion cues with explicit explanations, particularly for assistive navigation.

\subsection{Explainable Navigation}

% Explainable navigation aims to help users understand \emph{why} a robot moves as it does and, secondarily, anticipate what it will do next \cite{setchi2020explainable, dehkordi2021explainability, anjomshoae2019explainable}. Prior work formulates explanation generation as a design problem over modality, timing, and presentation \cite{halilovic2024planning, halilovic2024impact}, and recent systems leverage vision-language models to produce real-time navigation rationales, reporting improved user preference and understanding when transparency is provided during navigation \cite{sotomi2025trust}. Visualization-centric approaches further show that making navigation rationales explicit (e.g., via semantic projections in immersive interfaces) can improve comprehension and predictability for sighted users \cite{deheuvel2025immersive}, motivating alternative modalities in accessibility-focused settings. Explainable planning frames explanations as resolving mismatches between a user’s expectations and the robot’s internal model, enabling contrastive queries such as ``why did you deviate from going forward?'' \cite{chakraborti2017model}. For navigation, contrastive path explanations can be generated by identifying minimal changes that would make a user-expected path optimal, yielding actionable ``why A rather than B'' rationales \cite{brandao2021explaining}.
Explainable navigation aims to help users understand why a robot acts as it does and anticipate future behavior \cite{setchi2020explainable, dehkordi2021explainability}. Prior work formulates explanation generation as a design problem over modality, timing, and representation \cite{halilovic2024planning, halilovic2024impact}, while recent systems leverage vision-language models to generate real-time navigation rationales \cite{sotomi2025trust}. These approaches show that providing explanations can improve user preference and understanding during navigation. Other work frames explanation as resolving mismatches between user expectations and robot behavior, enabling contrastive reasoning such as explaining why one path was chosen over another \cite{chakraborti2017model, brandao2021explaining}. 
% Complementary approaches explore alternative modalities, including visualization-based explanations, which improve comprehension and predictability in controlled settings \cite{deheuvel2025immersive}.

% To support real-time, user-facing rationales in complex environments, recent work increasingly leverages foundation models to translate perceptual cues and navigation context into grounded natural-language explanations \cite{sotomi2025trust, song2024socialvlm, chen2024spatialvlm}. VLM-Social-Nav uses VLM prompting with detected entities to score navigation actions for socially aware planning \cite{song2024socialvlm}, while SpatialVLM couples an LLM with a spatially capable VLM for multi-step grounded spatial reasoning \cite{chen2024spatialvlm}. Building on these insights, we generate real-time motion explanations conditioned on image observations and the planner’s predicted future path to help users understand \emph{why} the robot acts a certain way and anticipate what it will do next. 
To support real-time explanations in complex environments, recent work increasingly leverages foundation models to translate perceptual inputs into grounded natural language \cite{song2024socialvlm, chen2024spatialvlm}. Building on these insights, we generate real-time motion explanations conditioned on image observations and predicted future trajectories to help users understand and anticipate robot behavior.

%\sehoon{We need 20 to 30 papers}
\section{Context-Aware Navigation Explanation Framework}\label{sec:method}
\begin{figure*}
    \centering
    \includegraphics[width=1\linewidth]{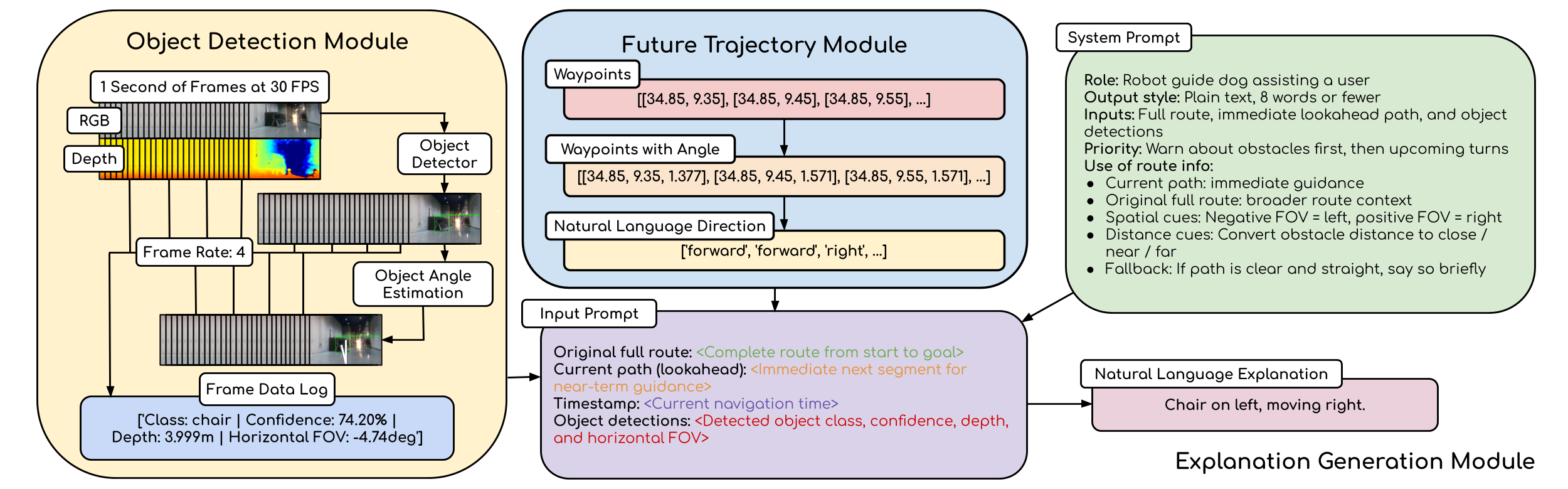}
    \caption{System overview of the Context-Aware Navigation Explanation framework. The object detection module extracts obstacle information from recent frames, while the future trajectory module provides planned waypoints and navigation intent. These inputs are combined into a structured prompt for an LLM to generate real-time natural language explanations.}
    \label{fig:system_overview}
    % \vspace{-1em}
\end{figure*}

% We formulate three modules to make up our system: object detection module, path planner module, and explanation generation module. 
Our goal is to generate concise but effective explanations of a robotic guide dog's navigation behavior during assistive navigation.
To achieve this, we propose a context-aware navigation explanation framework (Fig.~\ref{fig:system_overview}) that produces natural-language explanations for BVI users based on the robot's perception of nearby obstacles and its planned navigation trajectory.
These signals are combined to construct a navigation context, which is provided to an LLM to generate explanations of the robot's actions.
In the following sections, we describe how navigation context is represented as text, how explanations are generated using the LLM, and how explanation timing is controlled during navigation.

% Overview (screenshot as first draft)

% \subsection{Object Detection and Path Planner Module}\label{AA}
\subsection{Navigation Context Representation}
% The object detection and path planner modules gather frame and path data to input into the LLM. 
% To complete this task, the Intel RealSense D435i depth and RGB camera is used with YOLOv8n to detect objects and gather the depth and field of view for corresponding objects. 
% The path planner module collects planned path waypoints from the Go2 robot through the ROS2 platform. 
% This data collection occurs every five seconds and is translated from point cloud data to xy coordinate and yaw data for each waypoint in the form of (x, y, yaw).

% Navigation context overview: obstacle detection and traj.
To generate explanations grounded on the robot's behavior, the system constructs a navigation context that summarizes both the robot's perception of the environment and the resulting navigation trajectory toward the goal.
This representation captures information about nearby obstacles and the robot's intended motion, enabling the LLM to describe navigation decisions in a way that reflects the robot's current state.

% perception information-environmental info
% Environmental information is obtained from the robot's onboard RGB-D camera using an object detection model.
% For each detected object, the system records the object category, estimated distance from the robot, and its relative position within the robot's field of view. 
% These features provide a concise representation of obstacles that may influence the robot's navigation decisions. 
%\sehoon{be specific about the data type, like: At each time step, environmental information $e_t$ consist of X, Y, Z...}X

At each time step $t$, the system collects environmental information from the robot’s onboard RGB-D camera using an object detection model. For each detected object, the system records the object category, estimated distance from the robot obtained from depth sensing, and the relative horizontal position within the image frame. We denote the environmental information at time step $t$ as

\begin{equation}
e_t = \{(c_i, d_i, p_i)\}_{i=1}^{N_t},
\end{equation}

where $c_i$ denotes the detected object category, $d_i$ represents the estimated distance to the object, $p_i$ indicates the relative position in the robot's field of view, and $N_t$ is the number of detected objects at time step $t$.

% navigation trajectory
% In addition to perception data, the system incorporates information from the navigation planner.
% The planner produces a sequence of future waypoints that describe the robot's intended trajectory toward the goal point, where each waypoint is represented as $(x, y, \theta)$, specifying the robot's position and orientation in the local coordinate frame. 
% 
%\sehoon{define planner context.}X

In addition to perception data, the system incorporates information from a navigation planner. The planner produces a sequence of future waypoints describing the robot’s intended trajectory toward the goal point. Each waypoint is represented as $(x, y, \theta)$, corresponding to the robot's position and orientation in the global coordinate frame. We denote the planner context at time step $t$ as

\begin{equation}
p_t = \{(x_j, y_j, \theta_j)\}_{j=1}^{M},
\end{equation}

where $M$ is the number of future waypoints considered for explanation generation.

% context construction
% The detected objects and planned trajectory are combined to form the navigation context that is provided as input to the explanation generation module described in the following subsection. 
% \sehoon{Finally, define the navigation context as a union of environmental information and planner context. Be specific about time window.}

Finally, the navigation context is constructed by combining environmental observations over a short temporal window with the current planner trajectory. Let $W$ denote the number of perception frames aggregated for each explanation step. The navigation context at time $t$ is defined as

\begin{equation}
C_t = \{e_{t-W+1}, \dots, e_t\} \cup p_t.
\end{equation}

This combined representation captures both recent environmental observations and the robot’s planned trajectory, providing sufficient context for generating explanations of the robot’s navigation behavior.

% 
% \subsection{Explanation Generation Module}
\subsection{LLM-Based Explanation Generation}

The explanation generation module produces natural-language descriptions of the robot's navigation behavior based on the constructed navigation context. The navigation context $C_t$ is converted into a structured textual representation before being provided to the LLM. 
% 1b REVIEW

Specifically, the environmental observations and planner trajectory are serialized into text describing detected obstacles and upcoming navigation actions. This structured representation allows the LLM to interpret the robot's perception and motion plan in a human-readable format while preserving relevant spatial information.

% \begin{table}
%     \centering
%     \caption{System Prompts \courtney{TODO: Add prompts}}
%     \begin{tabularx}{\linewidth}{@{}cX@{}}
%     \toprule
%     \textbf{\shortstack{Prompt \\Condition}} & \textbf{\shortstack{Full\\Prompt}}\\ 
%     \midrule
%     C1  & ``"\\
%     C2, C3 & ``"\\
%     \bottomrule
%     \end{tabularx}
%     \label{tab:prompts}
% \end{table}

\begin{tcolorbox}[
    title=System Prompt,
    fonttitle=\bfseries,
    breakable,      % allows it to span multiple pages
    colback=gray!10,
    colframe=black,
    after={\captionof*{figure}{\textit{Note: Bold text included only in C2 and C3 (navigation context conditions). Sentence marked * included only in C1 (no navigation context).}}}
]
``You are a robot guide dog leading a user. Respond in 8 words or less. No markdown. No formatting. \textbf{You receive two path fields: `Original full route' (the complete route from start to goal, fixed — use for spatial awareness of what lies ahead) and `Current path (lookahead)' (immediate next few meters, use for urgent turn/distance guidance). Path segments use `forward Xm', `turn left', `turn right', joined by `then', ending with `goal'. Example: `turn right soon' or `turn left ahead'. Use `Original full route' for broader context.} Object detection: negative FOV is left of center, positive is right. *The objects detected are obstacles, so the user wants to move around them, not toward them. Estimate obstacle distance as close, near, or far. If path is clear and straight, say so briefly."
\end{tcolorbox}

To guide explanation generation, we design prompts
% (Table ~\ref{tab:prompts}) 
that instruct the LLM to summarize the robot’s navigation behavior based on the perception and trajectory information provided as a robotic guide dog assisting a visually impaired user. 
% 2a REVIEW
The LLM creates natural language explanations with more flexible descriptions instead of template explanations. This flexibility can also lead to conversation personalization in future work.  

% The prompt encourages concise descriptions that reference relevant obstacles or navigation actions when appropriate.

To ensure explanations remain easily interpretable, the generated responses are constrained to short phrases of at most eight words to encourage the model to produce brief navigation summaries such as obstacle avoidance actions, upcoming turns, or arrival events. 

An example of the structured navigation context and prompt used for explanation generation is shown in Fig.~\ref{fig:system_overview}.

\subsection{Explanation Timing and Context Aggregation}
% explanation generation interval
During navigation, explanations are generated periodically from the current navigation context.
Rather than generating explanations at every sensor update, the system aggregates perception and trajectory information over a short temporal window before providing it to the LLM to reduce redundant explanations and produce concise summaries of the robot's navigation behavior.

% context aggregation window
% The amount of information included in each explanation step is controlled by several parameters: the number of perception frames aggregated per LLM call (frame history) and the number of future waypoints from the navigation planner. 
% \sehoon{define hyper parameters}X
% Additionally, the frame rate of the perception systems determines how frequently new environmental observations are captured.
% These parameters determine the amount of navigation context provided to each LLM call.

The amount of information included in each explanation step is controlled by several parameters. The parameter $N_{\text{frames}}$ denotes the number of perception frames aggregated for each LLM call, while $f_{\text{fps}}$ represents the frame rate of the perception system. Additionally, the parameter $M$ determines the number of future waypoints included from the navigation planner.

% LLM call timing equation??
Based on these parameters, the interval between LLM calls is determined by the number of frames included in the frame history and the frame rate of the perception system.
The resulting LLM call interval is defined as the following:
\begin{equation}
    \Delta t_{\text{LLM}} = \frac{N_{\text{frames}}}{f_{\text{fps}},}
    \label{eq:llm_call_interval}
\end{equation}

% where $N_{\text{frames}}$ denotes the number of perception frames aggregated for each explanation, and $f_\text{fps}$ denotes the frame rate of the perception system. 
where $\Delta t_{\text{LLM}}$ represents the time interval between consecutive explanation generations.

Fig.~\ref{fig:system_overview} illustrates how perception frames are aggregated over time before triggering each LLM call.

The robot operates using the Unitree Go2 quadruped platform, which performs localization and path planning using a ROS2-based navigation stack. Given a target location on the map, the navigation planner computes a collision-free trajectory consisting of a sequence of waypoints. These waypoints are continuously updated during navigation and provided to the explanation framework as planner context.

The explanation generation framework runs concurrently with the navigation system, allowing the robot to produce real-time explanations of its navigation behavior while guiding the user toward the goal location.

\section{User Study}
% % Robot Platform
% We used a quadrupedal robot Go2 from Unitree to serve as our robot platform. The robot was outfitted with an off-the-shelf rigid harness around the middle of its body and a Bluetooth speaker which was placed on the top of its head.

% % Participants
% We recruited 13 participants with varied levels of sight ability. Our participants consisted of 3 BVI and 10 non-BVI. **Further demographic    

% \begin{itemize}
% % \subsection{Research Questions}
% %     \item Hypothesis: The quality and trustworthiness of the explanations is expected to increase based on the combination of frame and path data
    
% % \subsection{Participants}
% %     \item 13 Participants
% %     \begin{itemize}
% %         \item 3 bvi
% %         \item 10 non bvi
% %     \end{itemize}
    
% % \subsection{Procedure}
% %     \item Environment: CODA 15th floor hallway
    
% %     \item Predefined Paths a and b >> describe them and insert map photo/diagram
% %     \begin{itemize}
% %         \item 3 experiments with different parameters (table here)
% %     \end{itemize}

% \subsection{Measures}
%     \item Questionnaire
%     \begin{itemize}
%         \item Quality measurement (connection to a paper?): clear specific, relevance, preference** from Trust Through Transparency
%         \item Quality connection to trust (mapping )
%     \end{itemize}
% \end{itemize}

We conducted a user study to evaluate the effectiveness of LLM-generated explanations during robot-assisted navigation. The study examines how different configurations of navigation context influence the perceived quality of explanations and users' trust in the robotic guide dog system.

\subsection{Study Design}
We adopt a within-subject experimental design. We hypothesize that a richer navigation context will lead to higher explanation quality and improved user trust.
Specifically, explanations generated using both perception and trajectory information are expected to be perceived as clearer, more informative, and more trustworthy than explanations generated from a more limited context.

To examine this hypothesis, participants were exposed to multiple experimental conditions that varied the parameters used to construct the navigation context provided to the LLM (Anthropic's claude-haiku-4-5 model set with 50 maximum tokens on stream mode).
% 2a REVIEW

These conditions differed in the amount and type of navigation information included in each explanation step, as shown in Table~\ref{tab:condition}. 
C1 only provides obstacle information while C2 and C3 include obstacle and navigation information. Notifications for information such as turns and the end of the path are mentioned in addition to obstacles in C2 and C3. C3 also includes a longer frame history in order to not miss any obstacles which may be missed in C1 and C2. The frequency for all three conditions is also a key differentiator. C3 gives more information at a higher frequency, while C2 and C1 give less feedback.
% A summary of the experimental condition is provided in Table~\ref{tab:condition}.

\begin{table*}[ht]
    \centering
    \caption{User Study Conditions}
    \vspace{-.5em}
\begin{tabular}{ccccc}
\toprule
\textbf{Condition} & \textbf{\shortstack{Frame\\History}} & \textbf{\shortstack{Navigation\\Data}} & \textbf{\shortstack{LLM Call\\Interval(s)}} & \textbf{Example}  \\ \midrule
\textbf{C1}        &            1            &            No             &         3          &    ``Chairs detected left side, far away. Continue forward safely.''                   \\ 
\textbf{C2}        &            1           &              Yes           &          3        &         \shortstack{``Clear path ahead. Chair left 1.7m away. Continue forward.''}\\ 
\textbf{C3}        &            4            &              Yes           &          2        &        \shortstack{``Person left 4.7m. Continue forward, path clear ahead.''}                \\ \bottomrule
\end{tabular}
\label{tab:condition}
    \vspace{-.5em}
\end{table*}

\subsection{Participants}
A total of 12 participants were recruited for the study, including individuals with varied levels of visual ability to evaluate the effectiveness of the system across different user groups. 
Among the participants, two (B1 and B2) identified as blind or visually impaired (BVI), while the remaining 10 participants (S1 -- S10) were sighted.
% BVI primary mobility aid, other tool they tried, guide dog experience??
Sighted participants were asked to wear eye masks during the experiments to simulate reliance on the robot's guidance.
All participants provided informed consent prior to participating in the study.
The study protocol was reviewed and approved by the institutional review board (IRB).

\subsection{Experimental Setup}
% robot platform
The experiments were conducted using a quadrupedal robot platform, Unitree Go2, with an off-the-shelf rigid harness handle mounted around the center of its body.
A Bluetooth speaker was attached to the robot to deliver navigation explanations as spoken audio, generated via Google text-to-speech.
% software stack
The robot executed the navigation and explanation system described in Sec.~\ref{sec:method}.
We used an Intel RealSense D435i RGB-D camera with a YOLOv8n~\cite{jocher2023yolo} object detection model for perception information. 
The navigation planner Nav2~\cite{macenski2020nav2} collected planned waypoints through the ROS2 platform of the Go2 robot. 

\begin{figure}
    \centering
    \includegraphics[width=0.9\linewidth]{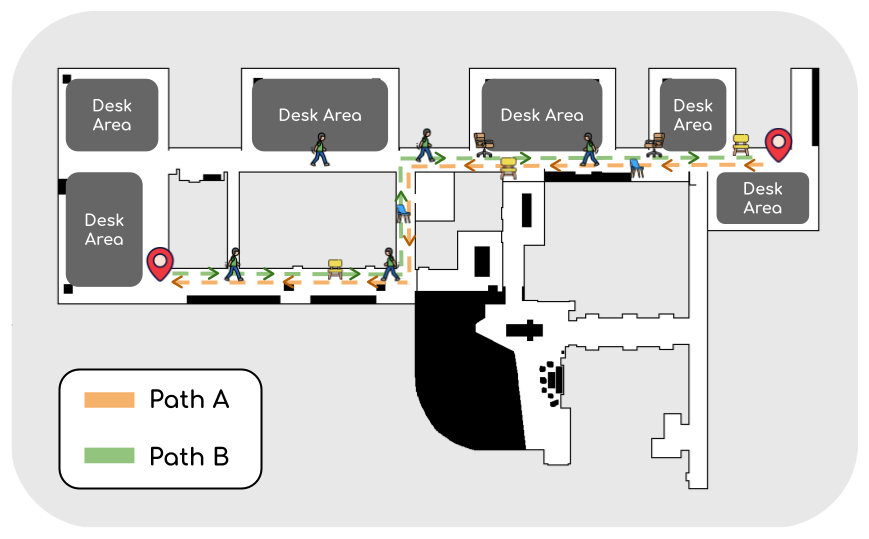}
    \vspace{-.5em}
    \caption{Experimental map showing two navigation routes (Path A and Path B), location of static obstacles (chairs), and dynamic agents (pedestrians) moving through the environment during evaluation of the explanation system.}
    \label{fig:map}
    \vspace{-1em}
\end{figure}

% environment/map
The study took place in an indoor hallway, which provides a structured navigation space with typical hallway obstacles (e.g., chairs, pedestrians) and turning points.
A predefined navigation path was used, where participants traversed in both directions of the path, denoted Path A and Path B in Fig.~\ref{fig:map}.
These paths were designed to include different obstacle layouts and navigation turns to trigger explanation events during assistive navigation.

\subsection{Procedure}
% nav trials and structure
At the beginning of the experiment, participants were introduced to the robotic guide dog system and the goal of the study.
Participants first complete a short familiarization phase to get comfortable following the robot's lead using the harness handle without navigation explanations.
% After this phase, participants were instructed to follow the robot while listening to the spoken navigation explanation generated during movement.
During each trial, the robot guided the participant along one of the predefined navigation paths while generating spoken navigation explanations.

Each participant completed six trials in total, across three experimental conditions and two paths.
The order of the conditions was counterbalanced across participants.
After each trial, participants complete a questionnaire evaluating the explanations they received during navigation.

\subsection{Measures} % this part will be updated after finalizing the questionnaires (probly after pilot study on Mon)
% metrics, questionnaire items, rating scale
% We evaluated the perceived quality of the generated navigation explanations using.

% We also evaluated participants' trust in the robotic guide dog system...

% All quantitative responses were collected using a Likert-scale questionnaire after each trial.
% ..

% qualitative???

%     \item Questionnaire
%     \begin{itemize}
%         \item Quality measurement (connection to a paper?): clear specific, relevance, preference** from Trust Through Transparency
%         \item Quality connection to trust (mapping )
%     \end{itemize}

After each trial, participants completed a post-run questionnaire consisting of trust-related items, explanation-related items, and an optional open-ended comment. 
All items were rated on a five-point Likert scale. 
% A summary of the questionnaire measures is provided in Table~\ref{tab:measures}.

To assess \textit{trust in the robot}, we included 14 items capturing participants' perceptions of the robot's reliability, predictability, dependability, responsiveness, and communication adapted from the Trust Perception Scale-Human Robot Interaction (TPS-HRI) \cite{schaefer2016tpshri}. 
%\taery{cite TPS}

To assess \textit{explanation quality}, we included seven items measuring how well the generated explanations helped participants understand the robot's navigation behavior. 
These items evaluated perceived understanding, satisfaction, detail, completeness, usefulness, and perceived informativeness of the explanations adapted from the Explanation Satisfaction Scale (ESS) \cite{hoffman2023ess}. 
%\taery{cite ESS}

Participants were also given the opportunity to provide open-ended comments after each run. 
These qualitative responses were used to contextualize the quantitative results and capture preferences or concerns not fully reflected by the Likert-scale ratings.

%\begin{table}
%    \centering
%    \caption{Summary of the post-trial questionnaire measures used in the user study, including the Robot Trust (TPS-HRI) and Explanation Satisfaction (ESS) scales, number of items, and representative example statements.}
%    \setlength\tabcolsep{2.5pt}
%\begin{tabular}{lcl}
%\toprule
%\textbf{Measure} & \textbf{\# Items} & \textbf{Example Items} \\ \midrule

%\begin{tabular}[t]{@{}l@{}}
%Robot Trust \\
%(TPS-HRI)~\cite{schaefer2016tpshri}
%\end{tabular} & 14 & 
%\begin{tabular}[t]{@{}l@{}}
%The robot is reliable \\
%The robot is predictable \\
%The robot follows directions
%\end{tabular} \\

%\begin{tabular}[t]{@{}l@{}} Explanation Satisfaction \\ (ESS)~\cite{hoffman2023ess} \end{tabular} & 7 & 
%\begin{tabular}[t]{@{}l@{}}
%I know how the robot navigates \\
%The explanation is satisfying \\
%The explanation has sufficient detail
%\end{tabular} \\

%\bottomrule
%\end{tabular}
%\label{tab:measures}
%\end{table}

\begin{figure}
    \centering
    \includegraphics[width=1\linewidth]{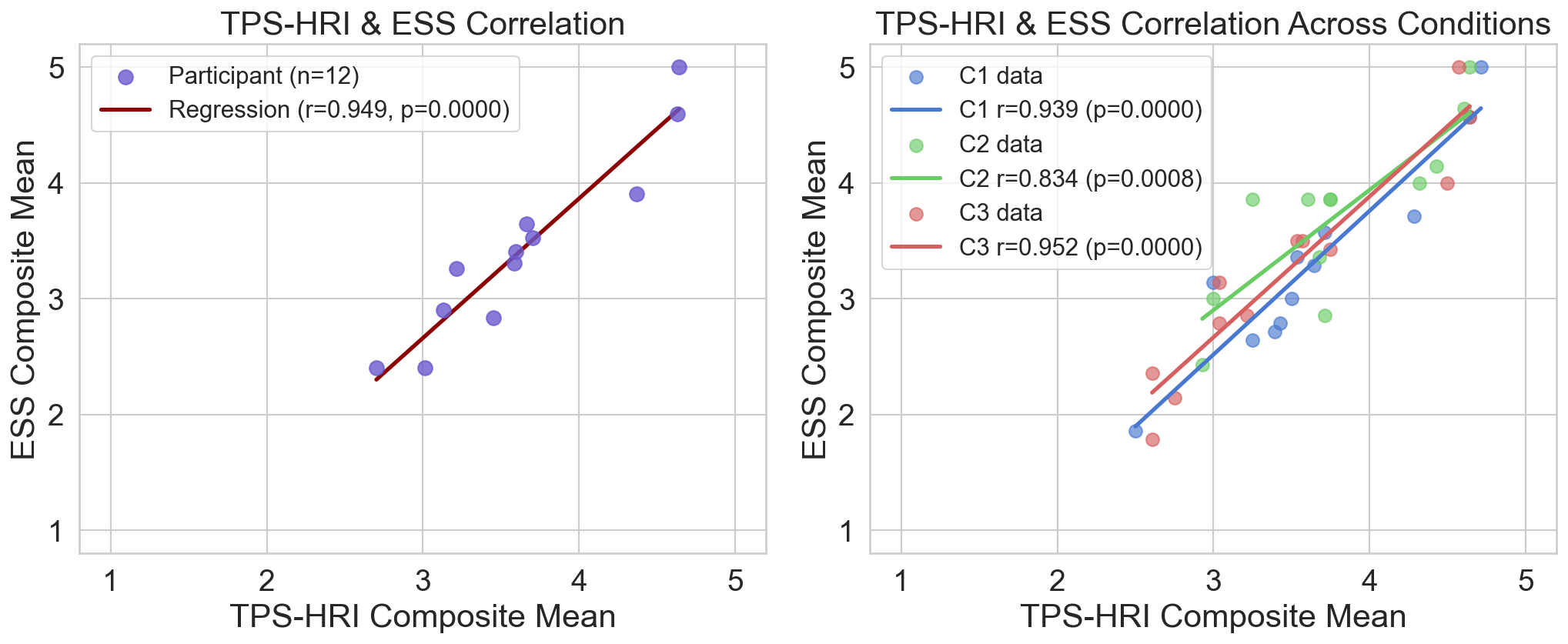}
    \caption{Trust and Explanation Satisfaction Correlation, which shows strong correlation between TRP-HRI across participants. \textbf{Left:} Aggregated plot. \textbf{Right:} Per condition plot.}
    \vspace{-1em}
    \label{fig:correlation}
\end{figure}

\section{Results}
% ratings analysis
We analyzed the relationship between robot trust (TPS-HRI) and explanation satisfaction (ESS) using correlation analysis and how these measures vary across our three conditions.

For analysis, we computed Pearson correlation coefficients to assess the correlation between TPS-HRI and ESS and used repeated-measures ANOVA to evaluate the effect of experimental conditions.
Significant effects were further examined using paired t-tests with Bonferroni correction for multiple comparisons.
For item-level analysis, we used the Friedman test to identify differences across conditions, followed by Dunn-Sidak corrected comparisons.

% H1: Higher ESS is positively correlated with higher TPS-HRI
% Pearson with Spearman as backup
% Fisher r to z tranformation showing correlation stability across conditions
% Summary of findings H1, H2, H3

\subsection{Participant-Level Analysis}
% All analyses were conducted at the participant level by averaging responses within each condition. \taery{please double check if this is correct}
In this section, we conducted the participant-level analysis to test three primary hypotheses on user trust and satisfaction:
\begin{itemize}
    \item H1: Participant trust in the robot is positively correlated with explanation satisfaction.
    \item H2: Trust scores differ significantly across our conditions, with navigation and obstacle information (C2 and C3) eliciting higher trust compared to obstacle information only (C1).
    \item H3: Explanation satisfaction scores differ significantly across our conditions, with navigation and obstacle information (C2 and C3) rated higher over obstacle information only (C1).
\end{itemize} 
\noindent\textbf{H1: Robot Trust and Explanation Satisfaction.}
We first examine the relationship between robot trust (TPS-HRI) and explanation satisfaction (ESS). 
Pearson correlation analysis revealed a strong positive association between TPS-HRI and ESS scores across participants $(r=0.949, p < 0.05)$, as shown in Fig.~\ref{fig:correlation}.
This result indicates that higher satisfaction with the robot's explanations is strongly associated with higher levels of trust in the robot, which supports \textbf{H1}.
To assess whether this relationship is consistent across different explanation conditions we computed correlation coefficients separately for each condition. 
As illustrated in Fig.~\ref{fig:correlation}, all conditions exhibit a similarly strong positive correlation, suggesting that the relationship between explanation satisfaction and trust is robust to variations in explanation design.

% Fisher to z? Explain what this means for our findings

%\begin{itemize}
%    \item H1
%    \begin{itemize}
%        \item Correlation calculation method
%        \item Correlation result and explanation
%        \item Correlation by condition calculation method
%        \item Correlation by condition result and explanation
%    \end{itemize}

%%%%%%%%%%%%%%%%%%%%%%%%%%%%%%%%%%%%%%%%%%%%%%%%%%%%%%%%%%%
% C1 & C2
% \textbf{Significant difference in trust and explanation satisfaction varies among conditions.} Repeated measures ANOVA showed significant differences in trust $(F(2,22)=3.951, p=0.0342)$ and explanation satisfaction $(F(2,22)=5.629, p=0.0106)$ across our conditions. 
% We conducted paired t-tests to compare significance across conditions. Participants rated explanation satisfaction significantly higher for C2 $(M=3.74, SD=0.73)$ compared to C1 $(M=3.30, SD=0.85), t(11)=2.644, p=0.023,$ Cohen's $d=0.763$. However, this comparison did not survive Bonferroni correction where $\alpha = 0.0167$ and $p_{\text{adj}}=0.0685$. Participants also rated C2 $(M=3.74, SD=0.73)$ significantly higher than C3 $(M=3.26, SD=0.96)$ which held for Bonferroni correction $(t(11)=2.98, p=0.013, p_{\text{adj}}=0.038,$ Cohen's $d=0.86)$. 

\begin{figure}
    \centering
        \includegraphics[width=0.8\linewidth]{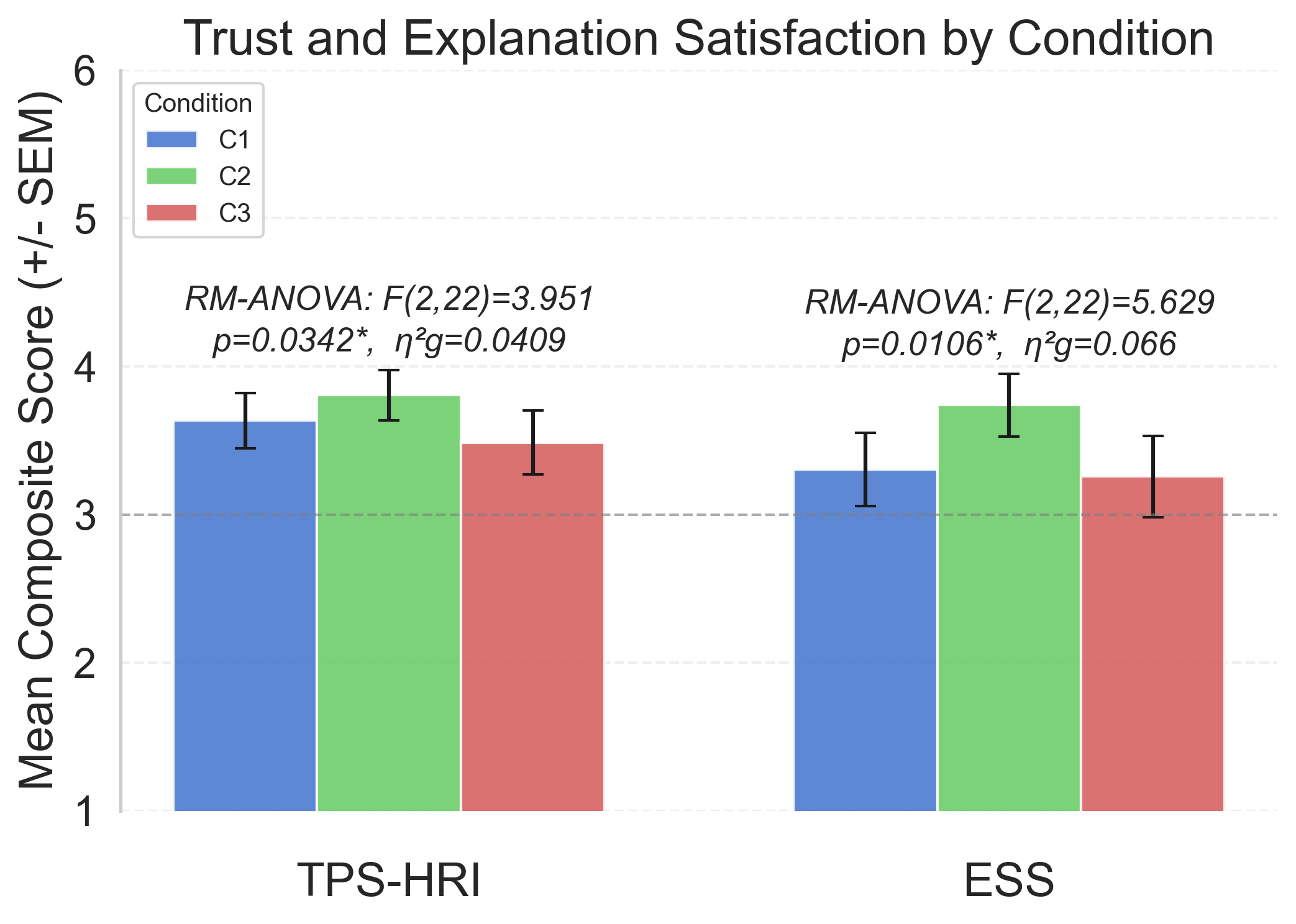}
        \vspace{-.5em}
        \caption{Mean scores for TPS-HRI \textbf{(left)} and ESS \textbf{(right)}. RM ANOVA shows significance across conditions.}
        \vspace{-1em}
        \label{fig:means}
\end{figure}

\noindent\textbf{H2 and H3: Effect of Conditions on ESS and TPS-HRI.}
We conducted repeated-measures ANOVA to examine the effect of the three conditions on TPS-HRI and ESS (Fig.~\ref{fig:means}). The results showed significant main effects of the explanation condition on both trust ($F(2, 22) = 3.951, p = 0.0342$) and satisfaction ($F(2, 22) = 5.629, p = 0.0106$). These findings demonstrate that the design of navigation explanations directly influences both perceived quality and user trust. Low effect size ($\eta^2_g = 0.0409$ for TPS and  $\eta^2_g = 0.066$ for ESS) is most likely due to a modest sample size (n=12).   

Post-hoc analyses using paired $t$-tests with Bonferroni correction provided further insights:
\begin{itemize}[noitemsep, topsep=0pt]
    \item \textbf{Robot Trust (TPS-HRI)}: Similar to satisfaction, C2 ($M=3.81, SD=0.59$) received higher trust ratings than C3 ($M=3.49, SD=0.75$; $t(11) = 2.70, p = 0.021, d = 0.78$), although the result did not survive Bonferroni correction ($p_{adj} = 0.063$). 
    \item \textbf{Explanation Satisfaction (ESS)}: C2 ($M=3.74, SD=0.73$) was rated significantly higher than C3 ($M=3.26, SD=0.96$; $t(11) = 2.98, p = 0.013, p_{adj} = 0.038, d = 0.86$). While C2 also outperformed C1 ($M=3.30, SD=0.85$), the difference was not statistically significant after correction ($t(11) = 2.644, p = 0.023, d = 0.763$).
\end{itemize}

Despite the marginal significance in trust after correction, the high effect sizes ($d = 0.78$ and $0.86$) suggest that providing integrated context at a moderate frequency (C2) is more effective than high-frequency updates (C3). These results partially support for \textbf{H2} and \textbf{H3}, identifying C2 as the optimal configuration for  user understanding and trust.

\subsection{Item-Level Analysis}
\begin{figure}[H]
    \centering
        \begin{subfigure}{\linewidth}
            \centering
            \includegraphics[width=\linewidth]{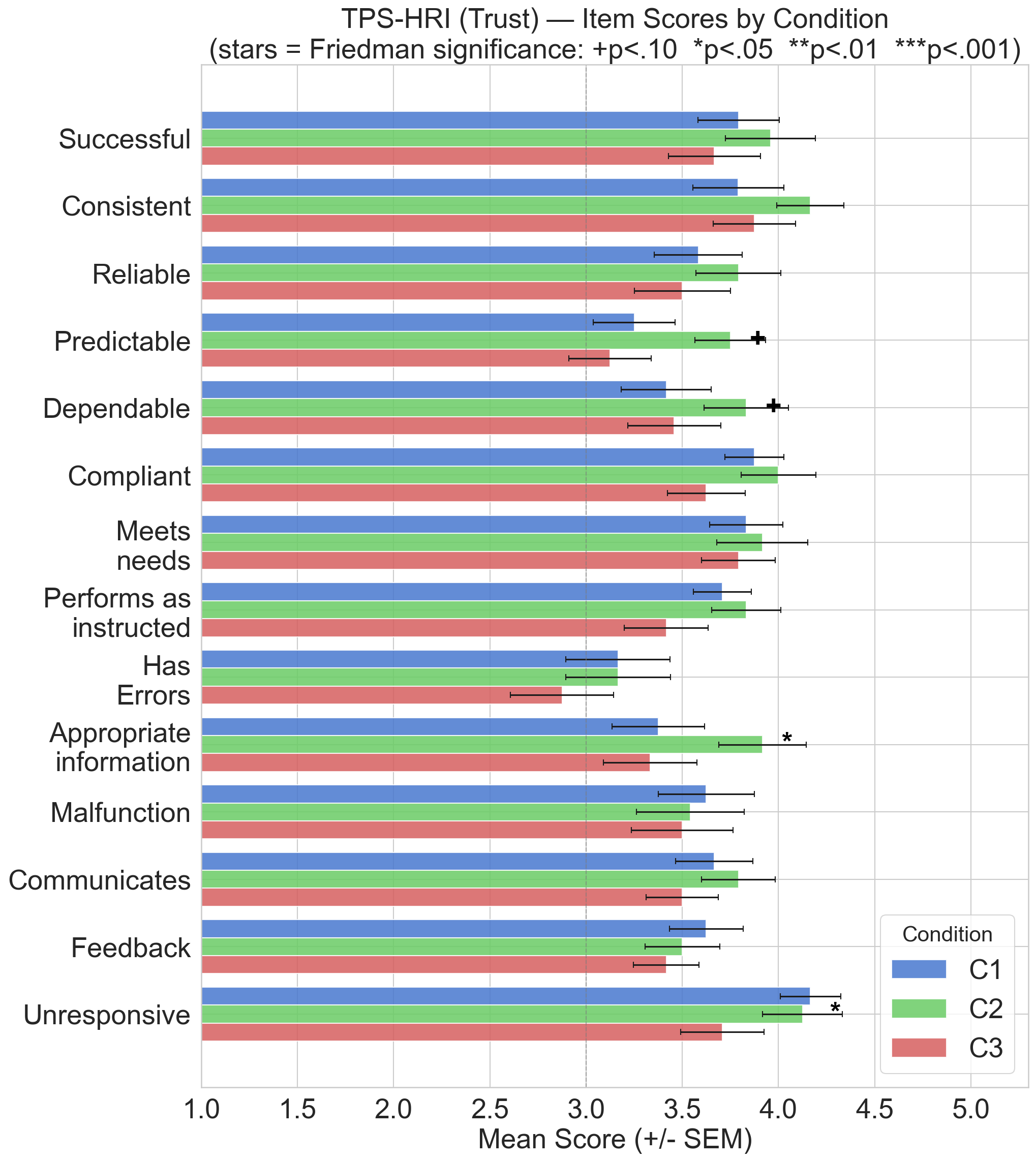}
            \caption{TPS-HRI}
            \label{fig:tps_items}
        \end{subfigure}
        \vspace{1em}
        \begin{subfigure}{\linewidth}
            \centering
            \includegraphics[width=\linewidth]{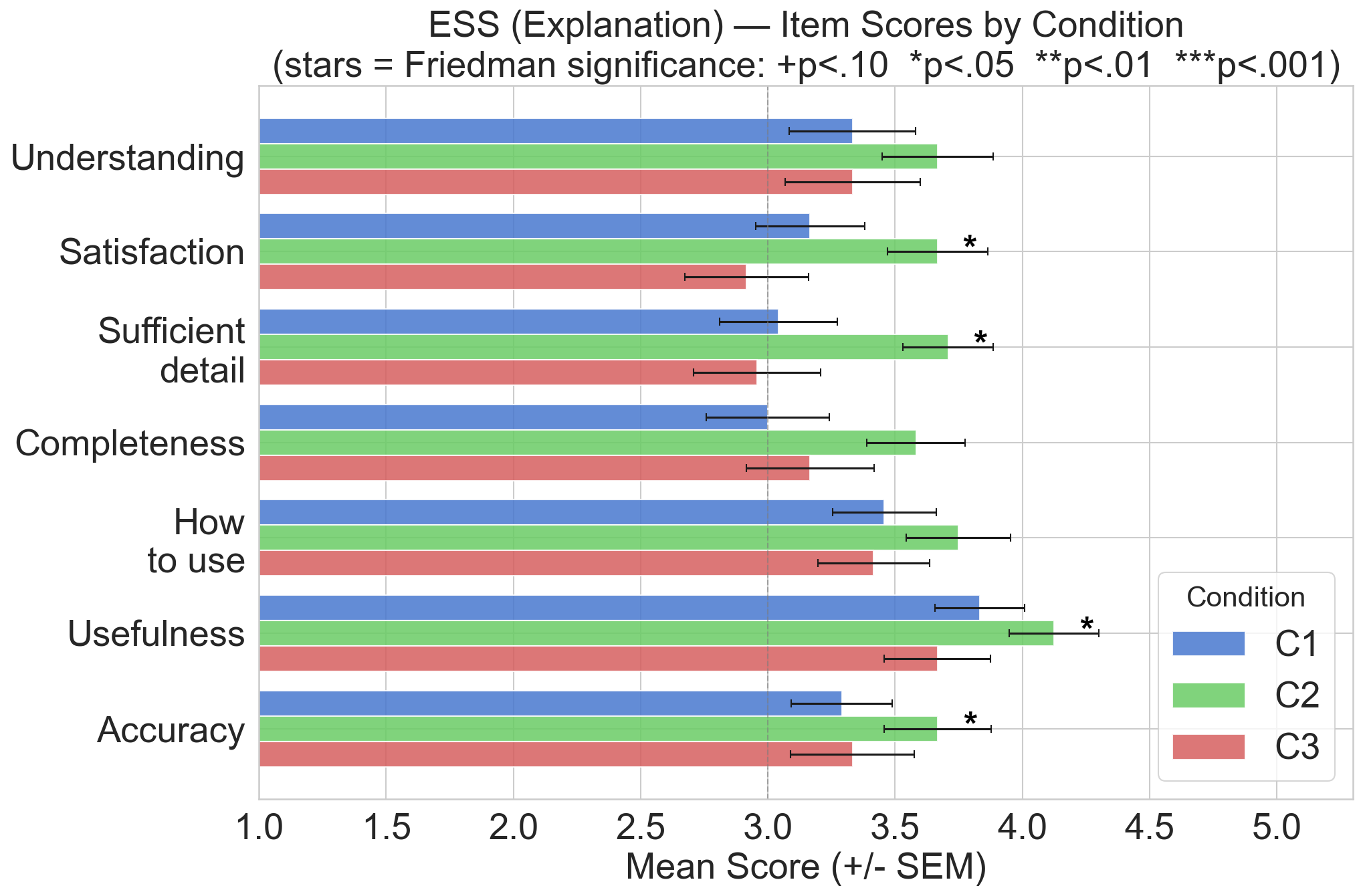}
            \caption{ESS}
            \label{fig:ess_items}
        \end{subfigure}
    \vspace{-2em}
    \caption{
        Item-level mean scores for \textbf{(a)} trust (TPS-HRI) and \textbf{(b)} explanation satisfaction (ESS) across conditions (C1-C3). The following items have significant p values below 0.05: appropriate information, unresponive, satisfaction, sufficient detail, usefulness, and accuracy.  
    }
\end{figure}

To better understand which aspects of explanations contribute to user satisfaction and trust, we conducted an item-level analysis across conditions (Fig.~\ref{fig:tps_items} and Fig.~\ref{fig:ess_items}).
A Friedman test revealed significant differences across conditions for both trust- and explanation-related items (p $<$ 0.05), indicating that specific aspects of the interaction vary with explanation design.

%\courtney{double check items}
Post-hoc comparisons using Dunn-Sidak show that items across TPS-HRI and ESS were consistently rated higher in C2 compared to other conditions.
For example, usefulness and satisfaction were rated significantly higher in C2 than C3 ($p = 0.016, p_{adj} = 0.046$).

Items such as sufficient detail ($p = 0.019, p_{adj} = 0.058$) and completeness ($p = 0.016, p_{adj} = 0.046$) showed more favorable ratings towards C2 over C1, suggesting that explanations providing richer and more actionable information are perceived more favorably. 

For trust, participants rated the robot as more dependable ($p = 0.047, p_{adj} = 0.134$) in C2 over C1 and as providing more appropriate information ($p = 0.031, p_{adj} = 0.091$) in C2 over all other conditions, indicating that clear and informative explanations contribute to perceived trust. 

Across both measures, C2 consistently received higher ratings than C1 and C3, suggesting that explanation designs that balance contextual detail and clarity are most effective in supporting user satisfaction and trust.

% Diagrams
% overall correlation

% all statistical results
% start with one-liner summary
% each hypothesis with stat result and why

% \\\\
% tps -- ess correlation
% \textbf{Explanation satisfaction is positively correlated with trust.} 
% We used the Pearson correlation coefficient to calculate correlation for explanation satisfaction and trust which showed significant results $(p<0.05)$. For a combined correlation across all conditions for each participant, we observe a positive, high correlation $(r=0.949, p < 0.05)$ in Fig.~\ref{fig:overall_corr}. Our claim is further strengthened by observing (Fig.~\ref{fig:cond_corr}), which signals high correlation for each condition, indicating that our claim is consistent across conditions.
\section{Discussion}
% General discussion
\subsection{General Discussion}
\noindent\textbf{Detection-action combination.} Participants preferred hearing the robot mention an obstacle and then give a directive or actionable comment on how to move around it. Users noted this combination of information helped them prepare for the robot's next movement. Additionally, when these movements were completed, participants reported increased trust in the robot due to it carrying out the aforementioned action. 

%\begin{itemize}
%    \item Participants prefer robot calls out obstacle and give directive on what to do to avoid obstacle 
%    \item on obstacle description: B1 "Gave me an idea of what a chair looks like in my head even if I didnt see it very well."
%    \item S4 ""
%\end{itemize}

%Participants also expressed satisfaction in hearing about an obstacle more than once; when the obstacle was further away and when it was closer to the user.  
%\begin{itemize}
%    \item too frequent on C3 for cognitive load for sighted participants but no issue on this from bvi, may be due to space unfamiliarity or more reliance on sense of hearing
%    \item participants enjoyed object mentioned far away and when getting near
%    \item B1 ""
%\end{itemize}

%\begin{itemize}
%    \item Some participants enjoy specific distance given S11 "I prefer if [the robot] tells me the exact distance" and S12 "I like the exact [amount] of distance...I think I can more clearly expect what [is going to happen]."
%    \item Distance not as helpful for bvi S1"For a visually impaired person or blind person distance is sometimes hard to measure, so having it said enough ahead of time as the robot did, that can alert the person to be cognizent that on the left , coming up there will be an obstacle"
%    \item some enjoyed general statements (far, near)
%    \item some thought feet or steps would be better 
%\end{itemize}
% BVI Participants
\subsection{BVI Participant Preferences and Suggestions}
\noindent\textbf{Distance awareness.} Our findings reveal a distinct divergence in spatial perception: while BVI users and the majority of participants favored general distance descriptions, a subset of sighted users preferred exact numerical metrics. BVI participants, such as B1, noted that precise distances are often difficult to conceptualize in real-time, explaining that qualitative alerts like ``near'' or ``coming up'' allowed them to be more cognizant of upcoming obstacles well in advance. Conversely, sighted participants like S9 and S10 expressed a preference for exact distances, with S10 noting that numerical data allowed them to ``more clearly expect'' the robot's next move. Additionally, some participants suggested that communicating distance in steps might offer a more intuitive alternative to standard units.
% \noindent\textbf{Distance awareness.} Some participants enjoyed hearing specific distances to the next obstacle or turn to help them understand what is happening next. S9 said, "I prefer if [the robot] tells me the exact distance," and S10 elaborated saying, "I like the exact [amount] of distance...I think I can more clearly expect what [is going to happen]." Other participants enjoyed more general statements (e.g. "far","near", "coming up") from the robot which were most consistent in C1 due to the lack of specific navigation information. This was mentioned by B1 who stated, "For a visually impaired person or blind person distance is sometimes hard to measure, so having it said enough ahead of time as the robot did, that can alert the person to be cognizant that on the left , coming up there will be an obstacle." Still, others preferred having distance information in steps.

\noindent\textbf{Cognitive load.} BVI participants did not report the cognitive overload observed in sighted users during C3, despite the high frequency of the robot’s comments. This suggests that for BVI users, the increased information density was not a burden but perhaps a necessity due to their greater unfamiliarity with the space. Unlike sighted participants, who found the frequency excessive, BVI participants likely required these frequent updates to maintain continuous spatial orientation in an unknown environment.
% \noindent\textbf{Speech frequency variations.}  Sighted participants felt that the frequency of the robot's comments in C3 induced cognitive overload. However, BVI participants did not mention this as an issue. This may have been due to their increased unfamiliarity with the space compared to sighted participants, which could cause them to not be uncomfortable when receiving information more frequently.

\noindent\textbf{Obstacle awareness.} BVI participants found obstacle identification to be important. On obstacle description, B1 stated that identification, ``gave me an idea of what a chair looks like in my head even if I didn't see it very well.'' B1 also mentioned that object size would also be helpful, saying,``A cone in the road would be helpful to get an image for that person to know if it's a big object, if it's a medium sized, or smaller object,'' and explained, ``For somebody that [is] totally blind I think that could be very helpful... to be as specific as [the robot] can be possible, so [the user] can get around whatever is coming up.'' B2 also indicated curiosity in upcoming obstacles,  saying ``When [the robot] says something to the right, I was kind of curious.'' and gestured reaching out to feel what the obstacle was. Additionally, BVI participants enjoyed hearing about an obstacle more than once; when the obstacle was further away and when it was closer to the user. 
% B1 would have also liked to know when an obstacle had been cleared, ""
%Some sighted participant (S5) did not find obstacle identification to be important to them, however 

\noindent\textbf{Human-robot team.} BVI participants viewed themselves and the robot guide dog as a team or as ``us'' as described by B2, which is the traditional concept behind human-guide dog teams. B1 stated, ``The robot seemed to want me to be safe,'' signaling trust in the robot as a partner on a team. B1 also suggested the robot dog to introduce itself, saying, ``...you become like a team and I think it's so much more personal if he identifies or she identifies her name... Even something that sounds maybe inane or very simple makes that person feel that they are a team.''
% B2 mentions our path, bc robot is guiding "us" R2

\noindent\textbf{Definitive language.} Both BVI participants mentioned the definitive nature of the robot's comments. These comments made the robot more trustworthy to them because it was not guessing or sounding unsure in its delivery. B1 stated that the robot was ``very definitive,'' and ``made me feel safe, secure with the comments and that the robot was going to implement them well, which he did.'' B2 similarly stated ``[The robot] knew what it was talking about, it wasn't guessing.'' This participant also compared the robot to a GPS, and said, ``It wasn't vague in it's description, which made me feel comfortable.'' 

%\begin{itemize}
    %\item object description (more for bvi, less for sighted) B1 "A cone in the road would be helpful to get an image for that person to know if it's a big object, if it's a medium sized, or smaller object" "Even if not by name, by size" example:"carriage with wheels" B1 "For somebody that [is] totally blind I think that could be very helpful" "to be as specific as it can be possible, so they can get around whatever is coming up"
    
%    \item cleared object B1 " 
%    \item we are starting, "we" for team (friend) B1 "The robot seemed to want me to be safe" "I think just introducing himself because you become like a team and I think it's so much more personal if he identifies or she identifies her name...Even something that sounds maybe inane or very simple makes that person feel that they are a team."
%    \item robot was definitive in comments (no guessing) B1"Very definitive...and that made me feel safe, secure with the comments and that the robot was going to implement them well, which he did."
%\end{itemize}
% Limitations
\subsection{Limitations}
In our setup, the robot does not have spatial awareness for the user, and YOLO did not detect walls or support beams that some participants experienced minor collisions with. This limitation may have affected participants' overall trust in the robot. However, since they were present across all experimental conditions, they likely affected conditions similarly, allowing us to examine the relative impacts.
% of explanation context on trust and satisfaction.
% which was not due to the robot's comments but rather a design choice. This will be investigated for future studies. 
The in-application 3D map used for the Go2 robot also experienced occasional localization errors at times due to the uniform design of the space. 
Future work could address this limitation by integrating multi-sensor fusion techniques to ensure robust navigation in feature-less environments.
Another limitation includes errors in explanations (specifically for C2 and C3), where the system incorrectly told the user there was a turn ahead. For example, S9 commented about a mismatch in robot navigation and explanation, lowering the participant's satisfaction in the comments. 
% REVIEW 3b
%\taery{any comments related to this? e.g., the explanation was wrong} 
% \begin{itemize}
%    \item Robot not knowing user is there (no spatial awareness for user) and YOLO doesnt detect walls and beams  so some collisions with walls/cement support beams
%    \item Indoor env localization 
%    \item Navigation error/LLM hallucinating turns for C1 dilute difference between C1 (no navigation waypoints) versus C2 and C3 (both include navigation waypoints)
%\end{itemize}

% general discussion
% highlight bvi participants
% limitations

% Use user id's
\section{Conclusion}
%\begin{itemize}
%    \item Summary of contributions
%    \item Future work
%    \begin{itemize}
%        \item Emergency alert
%        \item Haptic feedback ref Towards Robotic Companions: Understanding Handler-Guide
%Dog Interactions for Informed Guide Dog Robot Design and Understanding Expectations for a Robotic Guide
%Dog for Visually Impaired People (mentions haptic feedback in both)
%        \item Integration with safety aware RL/action shielding (Taery)
%    \end{itemize}
%\end{itemize}

In this work, we present a context-aware navigation explanation framework that generates real-time natural language explanations for a robotic guide dog using LLMs. By combining perception-based obstacle information with future navigation trajectory, the system provides concise explanations that reflect the robot’s behavior and intent.

Through a user study, we show that explanation satisfaction is strongly correlated with user trust, suggesting that improving explanation quality is critical for increasing trust in assistive robotic systems. We also find that combining obstacle information and future trajectory with a moderate explanation frequency tended to receive higher ratings in both explanation satisfaction and trust than conditions without future trajectory or higher frequency. Item-level analysis further indicates that users value explanations that are informative, complete, and useful to their goals.

These results highlight the importance of providing context-rich and actionable explanations during navigation, especially in assistive settings where users rely on the robot to understand the environment.

For future work, we plan to extend this system with additional modalities such as haptic feedback and emergency alerts to improve safety and accessibility. We also aim to integrate safety-aware control mechanisms, such as the action-shielding approach proposed by Kim et al. \cite{kim2023transforming}, to improve robustness and reliability. Finally, we plan to conduct studies with a larger BVI population to better understand long-term trust and usability in real-world environments.

\section*{Acknowledgement}
% This research has been funded by the Industrial Technology Innovation Program (P0028404, development of a product level humanoid mobile robot for medical assistance equipped with bidirectional
% customizable human-robot interaction, autonomous semantic navigation, and dual-arm complex manipulation capabilities using large-scale artificial intelligence models) of the Ministry of Industry,
% Trade and Energy of Korea.
This research has been funded by the Industrial Technology Innovation Program (P0028404) of the Ministry of Industry, Trade and Energy of Korea.

\bibliographystyle{IEEEtran}
\bibliography{refs}
%\input{manuscript/99-appendix}
% \courtney{taking out appendix, full prompt input into manuscript and participants could be added as table, full list not needed (seen in other roman papers)}
\end{document}